\documentclass[a4paper,12pt]{article}
\usepackage[utf8]{inputenc}
\usepackage[english]{babel}
\usepackage{authblk}
\usepackage{graphicx}
\usepackage{mathptmx}
\usepackage[singlespacing]{setspace}
\usepackage[headheight=1in,margin=1in]{geometry}
\usepackage{fancyhdr}
\usepackage{lipsum}
\usepackage{booktabs}
\usepackage{makecell}
\usepackage{graphicx}
\usepackage{multirow}

\makeatletter
\def\@maketitle{%
  \newpage

  \begin{center}%
  \let \footnote \thanks
    {\LARGE \@title \par}%
  \end{center}%
  \par
  \vskip 0.1em}
\makeatother
\graphicspath{{images/}}

\title{Optimizing LLM Annotation of Classroom Discourse through Multi-Agent Orchestration
}

\date{}

\begin{document}

\maketitle
\thispagestyle{fancy}
\noindent
{\large
BAKHTAWAR AHTISHAM, Cornell University, USA\\
KIRK VANACORE, Cornell University, USA\\
RENE F. KIZILCEC, Cornell University, USA
\par}
\begin{center}
\newline
\end{center}
Large language models (LLMs) are increasingly positioned as scalable tools for annotating educational data, including classroom discourse, interaction logs, and qualitative learning artifacts [2, 8,9,10]. Their ability to rapidly summarize instructional interactions and assign rubric-aligned labels has fueled optimism about reducing the cost and time associated with expert human annotation [1, 2]. However, growing evidence suggests that single-pass LLM outputs remain unreliable for high-stakes educational constructs that require contextual, pedagogical, or normative judgment, such as instructional intent or discourse moves [3, 4]. This tension between scale and validity sits at the core of contemporary education data science. In this work, we present and empirically evaluate a hierarchical, cost-aware orchestration framework for LLM-based annotation that improves reliability while explicitly modeling computational tradeoffs. Rather than treating annotation as a one-shot prediction problem, we conceptualize it as a multi-stage epistemic process comprising (1) an unverified single-pass annotation stage, in which models independently assign labels based on the rubric; (2) a self-verification stage, in which each model audits its own output against rubric definitions and revises its label if inconsistencies are detected; and (3) a disagreement-centric adjudication stage, in which an independent adjudicator model examines the verified labels and justifications and determines a final label in accordance with the rubric. This structure mirrors established human annotation workflows in educational research, where initial coding is followed by self-checking and expert resolution of disagreements [5, 6].
\section*{Data and Task}

We evaluate our framework on a classroom discourse annotation task grounded in authentic K--12 mathematics instruction. The dataset comprises 63 classroom transcripts spanning whole-class discussion, small-group work, and online lessons. Teacher utterances were annotated by expert educators using the Talk Moves coding scheme \cite{talkmoves}, achieving high inter-rater reliability (Cohen's $\kappa > 0.90$). To balance construct coverage with computational feasibility, we applied proportional stratified sampling to select 800 target teacher utterances, preserving the empirical distribution of Talk Moves categories. These utterances were embedded within their surrounding conversational context, yielding 467 dialogue segments containing over 13,000 utterances. Only teacher turns were evaluated; student turns were included solely as contextual input.

We evaluate six annotation strategies organized into two pipelines. The first pipeline uses standard language models and includes \textit{single-pass annotation}, \textit{self-verified annotation}, and \textit{adjudicated annotation}. The second pipeline uses reasoning-enabled models and includes \textit{reasoning-annotated}, \textit{reasoning-verified}, and \textit{reasoning-adjudicated} strategies. These strategies are tested across six production LLM variants from three model families---GPT, Claude, and Gemini---including both reasoning-enabled models (GPT-5.2 Thinking, Claude 4.5 Opus, Gemini 3 Pro) and non-reasoning models (GPT-4.1, Claude 4.5 Sonnet, Gemini 3 Flash). This design allows us to disentangle the effects of model capability and orchestration structure. For each strategy, we additionally track prompt tokens, completion tokens, and total token usage to enable cost-aware comparisons across pipelines.

\section*{Results}
\begin{figure}
    \centering
    \includegraphics[width=1\linewidth]{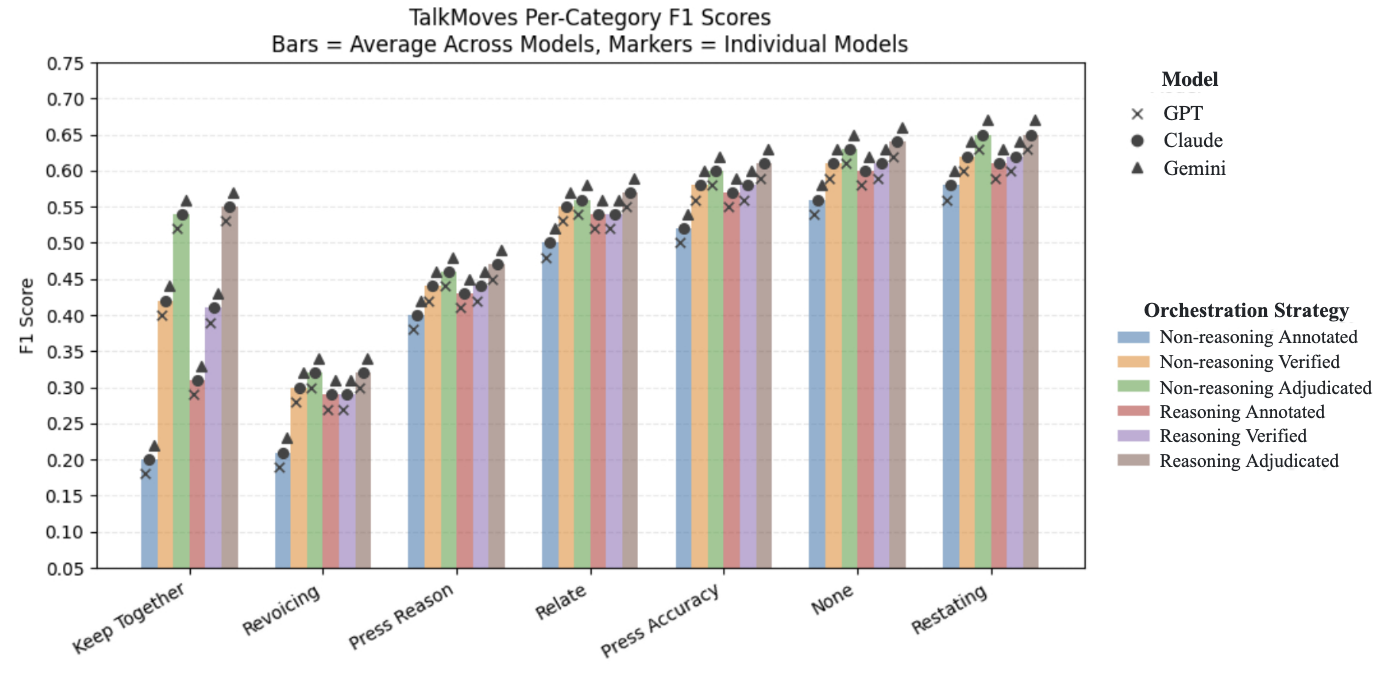}
    \caption{TalkMoves Per-Category Performance Across Orchestration Strategies. Average F1 scores for seven TalkMoves categories under different annotation orchestration strategies. Bars show mean performance averaged across three LLMs (GPT, Claude, Gemini), while markers indicate individual model scores (shapes). Categories are ordered from left to right by increasing baseline difficulty.}
    \label{fig:fig1}
\end{figure}
\textbf{Per-Category Performance Gains.} Fig. \ref{fig:fig1} presents per-category F1 scores for seven Talk Moves categories across all orchestration strategies, averaged across models, with markers indicating individual model performance. Across all categories, hierarchical orchestration yields consistent improvements over single-pass annotation. These gains are largest for pedagogically demanding categories that require interpretive or normative judgment. For example, \textit{Revoicing} improves from an average F1 of approximately 0.30 under single-pass annotation to 0.44--0.46 under verification and 0.52--0.54 under adjudication, corresponding to a $\sim$70--80\% relative improvement over the baseline. Similarly, \textit{Press for Reason} increases from $\sim$0.40 to $\sim$0.55--0.58 and $\sim$0.60--0.63, representing a $\sim$45--55\% relative gain (Fig.\ref{fig:fig1}; Table~1). In contrast, structurally simpler categories such as \textit{Restating} exhibit smaller but still systematic improvements, suggesting that orchestration confers benefits even when surface-level linguistic cues are more readily available. Together, these results indicate that hierarchical orchestration benefits categories with higher conceptual and pedagogical complexity more, while offering consistent gains across the full task spectrum.

\begin{figure}
    \centering
    \includegraphics[width=0.7\linewidth]{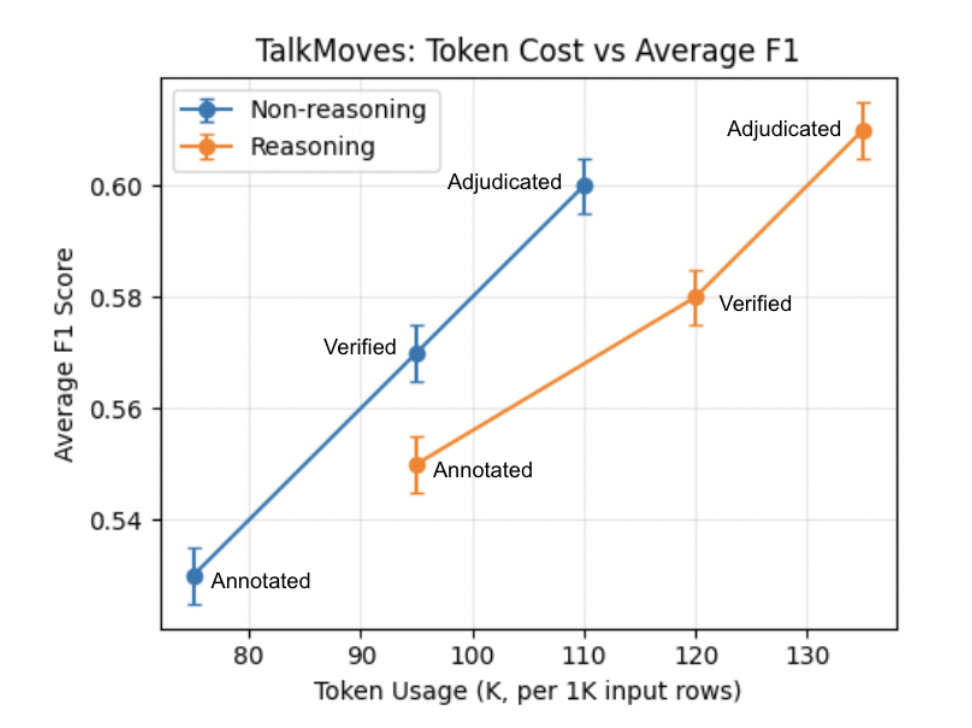}
    \caption{TalkMoves Token Cost vs. Average Performance for Reasoning and Non-Reasoning Pipelines. Average F1 score as a function of total token usage for TalkMoves annotation pipelines. Points trace the progression from single-pass annotation to verification and adjudication for non-reasoning and reasoning-based strategies.
}
    \label{fig:fig2}
\end{figure}

\textbf{Reasoning vs.\ Non-Reasoning Tradeoffs.} Reasoning-enabled models exhibit higher baseline performance under single-pass annotation. However, this advantage narrows substantially under orchestration. Figure \ref{fig:fig2} plots the average F1 against total token usage for Talk Moves pipelines. While reasoning-based adjudication achieves the highest absolute performance ($\sim$0.61 F1), non-reasoning adjudication reaches nearly the same level ($\sim$0.60 F1) at substantially lower token cost ($\sim$20--25\% fewer tokens). This reveals a clear Pareto tradeoff: reasoning improves initial annotation quality, but once verification and adjudication are introduced, orchestration structure---not model reasoning alone---becomes the dominant driver of reliability. Across multiple categories, non-reasoning models paired with adjudication match or outperform reasoning-enabled models with verification alone, despite lower per-call complexity. These results suggest that, for large-scale educational annotation pipelines, investing in orchestration---rather than relying exclusively on more expensive reasoning-capable models---can yield comparable reliability at lower cost. Taken together, the findings position hierarchical orchestration as a practical optimization strategy for scaling trustworthy LLM-based annotation in education data science.
\begin{table}[h]
\centering
\small
\resizebox{\textwidth}{!}{
\begin{tabular}{llcccccc}
\hline
Category & Model &
\makecell{Non-Reasoning \\ Annotated} &
\makecell{Non-Reasoning \\ Verified} &
\makecell{Non-Reasoning \\ Adjudicated} &
\makecell{Reasoning \\ Annotated} &
\makecell{Reasoning \\ Verified} &
\makecell{Reasoning \\ Adjudicated} \\
\hline

\multirow{3}{*}{Keep Together}
& Gemini & 0.22 & 0.44 & \textbf{0.56} & 0.33 & 0.44 & \textbf{0.57} \\
& GPT & 0.20 & 0.41 & \textbf{0.52} & 0.30 & 0.40 & \textbf{0.53} \\
& Claude & 0.19 & 0.42 & \textbf{0.53} & 0.31 & 0.41 & \textbf{0.54} \\
\hline

\multirow{3}{*}{Revoicing}
& Gemini & 0.21 & 0.32 & \textbf{0.34} & 0.31 & 0.32 & \textbf{0.35} \\
& GPT & 0.21 & 0.29 & \textbf{0.30} & 0.28 & 0.28 & \textbf{0.31} \\
& Claude & 0.20 & 0.30 & \textbf{0.31} & 0.29 & 0.29 & \textbf{0.32} \\
\hline

\multirow{3}{*}{Press Reason}
& Gemini & 0.42 & 0.46 & \textbf{0.48} & 0.45 & 0.46 & \textbf{0.49} \\
& GPT & 0.38 & 0.43 & \textbf{0.44} & 0.42 & 0.42 & \textbf{0.45} \\
& Claude & 0.40 & 0.44 & \textbf{0.46} & 0.43 & 0.44 & \textbf{0.47} \\
\hline

\multirow{3}{*}{Relate}
& Gemini & 0.52 & 0.56 & \textbf{0.58} & 0.55 & 0.56 & \textbf{0.59} \\
& GPT & 0.48 & 0.53 & \textbf{0.54} & 0.52 & 0.52 & \textbf{0.55} \\
& Claude & 0.50 & 0.55 & \textbf{0.56} & 0.54 & 0.54 & \textbf{0.57} \\
\hline

\multirow{3}{*}{Press Accuracy}
& Gemini & 0.54 & 0.60 & \textbf{0.62} & 0.59 & 0.60 & \textbf{0.63} \\
& GPT & 0.50 & 0.56 & \textbf{0.57} & 0.55 & 0.55 & \textbf{0.58} \\
& Claude & 0.52 & 0.58 & \textbf{0.60} & 0.57 & 0.58 & \textbf{0.61} \\
\hline

\multirow{3}{*}{None}
& Gemini & 0.58 & 0.63 & \textbf{0.65} & 0.62 & 0.63 & \textbf{0.67} \\
& GPT & 0.54 & 0.60 & \textbf{0.61} & 0.58 & 0.59 & \textbf{0.62} \\
& Claude & 0.56 & 0.61 & \textbf{0.63} & 0.60 & 0.61 & \textbf{0.64} \\
\hline

\multirow{3}{*}{Restating}
& Gemini & 0.60 & 0.64 & \textbf{0.66} & 0.63 & 0.64 & \textbf{0.67} \\
& GPT & 0.56 & 0.61 & \textbf{0.62} & 0.60 & 0.60 & \textbf{0.63} \\
& Claude & 0.58 & 0.62 & \textbf{0.64} & 0.61 & 0.62 & \textbf{0.65} \\
\hline

\end{tabular}}
\caption{F1 Scores for TalkMoves Categories by Model and Orchestration Strategy. Bold values indicate the highest F1 scores within each category, demonstrating that multi-model adjudication consistently provides better alignment with the ground truth.
}
\end{table}

\section*{Implications for Education Data Science}

These findings have three implications for education data science. First, evaluating LLMs based on single-pass accuracy substantially underestimates their attainable reliability when structured verification and adjudication are introduced. Annotation quality is not solely a function of model capability, but of the evaluative processes that surround model outputs. Second, hierarchical orchestration enables non-reasoning models to approach or, in some cases, match reasoning-level performance at markedly lower computational cost, a consideration that is especially important for large-scale deployments and resource-constrained educational contexts. Third, cost-aware analysis reveals diminishing returns from reasoning-based adjudication, indicating that verification-first pipelines often provide the most effective balance between validity and efficiency. More broadly, this work reframes LLM-enabled annotation as a measurement problem rather than a prediction task. By making tradeoffs between accuracy, cost, and uncertainty explicit, our framework advances a transparent and reproducible approach to educational analytics that aligns with emerging calls to move beyond demonstrations of technical possibility toward trustworthy, evidence-based use of AI in educational research and practice.
\begin{flushleft}
\setlength{\parindent}{0pt}
\setlength{\leftskip}{0pt}
\setlength{\hangindent}{1.5em}
\section*{References}
[1] Bommasani, R., et al. (2021). On the opportunities and risks of foundation models.
Stanford Center for Research on Foundation Models.

[2] Gilardi, F., Alizadeh, M., \& Kubli, M. (2023). ChatGPT outperforms crowd workers for text-annotation tasks.
Proceedings of the National Academy of Sciences, 120(30), e2305016120.

[3] Holmes, W., Bialik, M., \& Fadel, C. (2019). Artificial intelligence in education:
Promises and implications for teaching and learning. Center for Curriculum Redesign.

[4] Kizilcec, R. F., et al. (2023). Algorithmic fairness in education.
AERA Open.

[5] Krippendorff, K. (2018). Content analysis: An introduction to its methodology.
SAGE Publications.

[6] Cohen, J. (1960). A coefficient of agreement for nominal scales.
Educational and Psychological Measurement, 20(1), 37--46.

[7] Suresh, H., et al. (2022). The TalkMoves dataset.
In Proceedings of EMNLP.

[8] Liu, X., Zambrano, A. F., Baker, R. S., et al. (2025). Qualitative coding with GPT-4:
Where it works better. Journal of Learning Analytics, 12(1), 169--185.

[9] Meyer, J., et al. (2024). Using LLMs to bring evidence-based feedback into the classroom.
Computers and Education: Artificial Intelligence.

[10] Zhang, X., et al. (2024). A systematic literature review of empirical research on applying generative artificial intelligence in education.
Frontiers of Digital Education, 1(1), 1--24.

\end{flushleft}

\end{document}